\documentclass{llncs}

\usepackage[utf8]{inputenc}
\usepackage{graphicx}
\usepackage{amsmath}
\usepackage[font=footnotesize]{caption}
\usepackage{url}

\usepackage{caption}
\usepackage{subcaption}
\usepackage{booktabs}
\usepackage[table]{xcolor} 
\usepackage{xcolor}

\usepackage[breaklinks=true,colorlinks,bookmarks=false,linkcolor=blue]{hyperref}

\usepackage[sort&compress,capitalize,nameinlink]{cleveref}
\creflabelformat{equation}{#2#1#3}
\crefname{figure}{Fig.}{Figs.}

\begin{document}

\pagestyle{headings}  

\title{Beyond Averages: Open-Vocabulary 3D Scene Understanding with Gaussian Splatting and Bag of Embeddings}
\titlerunning{short title}  
\author{Abdalla Arafa\inst{1} \and Didier Stricker\inst{2}}
\authorrunning{Abdalla et al.} 
\institute{\email{abdalla.arafa@dfki.de}
\and
\email{didier.stricker@dfki.de}}

\maketitle              

\begin{abstract}
Novel view synthesis has seen significant advancements with 3D Gaussian Splatting (3DGS), enabling real-time photorealistic rendering. However, the inherent fuzziness of Gaussian Splatting presents challenges for 3D scene understanding, restricting its broader applications in AR/VR and robotics. While recent works attempt to learn semantics via 2D foundation model distillation, they inherit fundamental limitations: alpha blending averages semantics across objects, making 3D-level understanding impossible. We propose a paradigm-shifting alternative that bypasses differentiable rendering for semantics entirely. Our key insight is to leverage predecomposed object-level Gaussians and represent each object through multiview CLIP feature aggregation, creating comprehensive "bags of embeddings" that holistically describe objects. This allows: (1) accurate open-vocabulary object retrieval by comparing text queries to object-level (not Gaussian-level) embeddings, and (2) seamless task adaptation: propagating object IDs to pixels for 2D segmentation or to Gaussians for 3D extraction. Experiments demonstrate that our method effectively overcomes the challenges of 3D open-vocabulary object extraction while remaining comparable to state-of-the-art performance in 2D open-vocabulary segmentation, ensuring minimal compromise.
\end{abstract}

\section{Introduction}

Novel view synthesis has gained significant popularity since the introduction of NeRF\cite{nerf} (Neural Radiance Fields) in 2020, revolutionizing the field with its ability to render photorealistic scenes from novel viewpoints using neural networks. While NeRF pioneered neural scene representations with implicit volumetric rendering, recent advances such as 3D Gaussian Splatting\cite{gaussian_splatting} (3DGS) have shifted the paradigm by introducing efficient, explicit representations. By modeling scenes as collections of anisotropic 3D Gaussians, 3DGS achieves high-fidelity real-time rendering, unlocking new possibilities for dynamic scene reconstruction\cite{4d_gs_iso_rot,4d_gs_rotor,deformable_gs,urban_gs}, photo-realistic human avatars\cite{gaussian_avatars,3dgs_avatar},  and text-to-3D generation, where 3DGS serves as a differentiable intermediate representation for 2D diffusion models\cite{gaussian_dreamer,GSGen,dreamfusion}.

3D Gaussian Splatting achieves remarkable performance in novel view synthesis, generating photorealistic renderings in real-time. However, its fuzzy, unstructured nature presents a key limitation: the inability to clearly distinguish or segment individual objects within the scene. Unlike explicit mesh-based representations, the splatted Gaussians lack inherent semantic or geometric boundaries, making tasks such as object-level editing, scene understanding, or compositional manipulation challenging. To address these limitations, recent works\cite{langsplat,LEGaussians,feature3dgs} have proposed distilling semantic knowledge from 2D foundation models such as CLIP\cite{clip} and DINOv2\cite{dino_v2} with the help of extracted masks from SAM\cite{sam} into 3D Gaussians using differentiable rendering. Although these methods\cite{langsplat,LEGaussians,feature3dgs} correctly aggregate semantics in 2D image space - allowing open-vocabulary segmentation for rendered views—they struggle to learn clean object-level semantics at the individual Gaussian level. The noisy, inconsistent semantic attributes learned by the 3D Gaussians prevent reliable 3D scene understanding. Consequently, such approaches are limited to 2D search tasks (e.g., segmenting objects in a rendered frame) and fail to support 3D semantic queries (e.g., retrieving all instances of a specific object in the scene). 

To overcome these limitations, we propose a fundamentally different approach that completely bypasses differentiable rendering for semantic learning. Instead of distilling noisy 2D semantics into individual Gaussians, we use a pre-trained 3D Gaussian scene (decomposed into distinct objects via Gaussian Grouping\cite{gaussian_grouping}) and extract multiview CLIP embeddings for each object, forming a comprehensive ‘bag of embeddings’ that fully represents its visual appearance. For open-vocabulary querying, we perform retrieval directly at the object level: given a text prompt, we match it against the extracted embedding bags to identify relevant objects, then propagate their IDs to downstream tasks enabling precise 2D segmentation (filtering pixels by object ID) or 3D extraction (selecting Gaussians by ID). Our method decouples rendering from semantics, avoiding the pitfalls of alpha-blended semantic dilution while maintaining compatibility with real-time Gaussian splatting pipelines.

The contributions of this work are summarized as follows:

\begin{itemize}
    \item \textbf{Object-level Semantic Encoding}: We propose to represent 3D Gaussian scenes as bags of multiview CLIP embeddings per object, eliminating the need for error-prone differentiable rendering of semantics.

    \item \textbf{Open-Vocabulary 3D Search by Design}: By operating on pre-grouped multiview scenes, our approach enables direct object-level retrieval for text queries, resolving the inconsistency of per-Gaussian semantics in prior work.

    \item \textbf{Task-Agnostic Semantic Propagation}: We demonstrate how object IDs can be seamlessly propagated to both 2D segmentation (pixel-level filtering) and 3D extraction (Gaussian selection), unifying open-vocabulary tasks without retraining.

\end{itemize}

\section{Related Work}

\subsection{Neural Rendering}

Neural rendering has emerged as a powerful approach to synthesizing photorealistic images by combining the principles of computer graphics with deep learning. Early work, such as Neural Radiance Fields (NeRF)\cite{nerf}, introduced implicit volumetric representations, enabling novel view synthesis of complex 3D scenes. Subsequent improvements have accelerated training\cite{instantngp}.

In addition to implicit representations, explicit approaches such as Gaussian splatting\cite{gaussian_splatting} have gained prominence. These methods use point-based representations with Gaussian kernels, offering efficient rendering and enhanced scalability for real-time applications. Such explicit methods are particularly useful in scenarios requiring interactive performance without sacrificing quality.

Using both implicit and explicit representations, neural rendering has significantly expanded the possibilities for rendering and understanding 3D scenes, making it a critical tool for applications in virtual reality, robotics, and beyond.

\subsection{Open Vocabulary Scene Understanding}

Recent advances in open-vocabulary scene understanding aim to integrate 2D vision-language foundation models (VLMs) with 3D scene representations, combining the semantic richness of models like CLIP\cite{clip} and SAM\cite{sam} with the geometric fidelity of Neural Radiance Fields (NeRF)\cite{nerf} and 3D Gaussian Splatting (3DGS)\cite{gaussian_splatting}. Language Embedded Radiance Fields (LERF)\cite{lerf} pioneered this integration by embedding CLIP features into NeRF, enabling spatial queries such as locating objects described in natural language (e.g., "blue mug"). However, LERF's reliance on volumetric rendering incurs significant computational costs and suffers from semantic inconsistencies due to varying CLIP embeddings across viewpoints. CLIP-NeRF\cite{clip_nerf} extends these capabilities to enable object editing but inherits the limitations of scalability and rendering efficiency inherent to NeRF.

The emergence of 3DGS as a rasterization-based alternative to NeRF has enabled faster, more efficient fusion of VLMs with 3D scene representations. LangSplat\cite{langsplat} uses CLIP embeddings and a scene-level autoencoder to achieve real-time open-vocabulary querying while resolving ambiguities like distinguishing multiple instances of the same object. Gaussian Grouping\cite{gaussian_grouping} combines SAM's zero-shot segmentation with multi-view mask aggregation to achieve open-vocabulary instance segmentation, enabling object-level grouping without predefined categories. Feature3DGS\cite{feature3dgs} distills features from 2D models like DINOv2\cite{dino_v2} or SAM into 3D Gaussians, facilitating semantic segmentation and high-quality rendering, although its static features limit dynamic language-driven interactions. LeGaussians\cite{LEGaussians} further pushes the boundary by jointly optimizing 3D Gaussians with CLIP embeddings through feature reconstruction losses, balancing semantic and geometric fidelity but facing challenges with feature drift due to viewpoint inconsistencies.

While these methods show significant progress, they mainly focus on pixel-level open-vocabulary understanding, which limits their ability to fully capture the semantics of 3D scenes. By relying on per-pixel features or aggregated masks, these approaches often struggle to distinguish individual objects or entities in 3D space. This makes it difficult to handle tasks that require explicit object-level reasoning, such as understanding spatial relationships or interactions in 3D environments. Overcoming this limitation is essential for advancing open-vocabulary scene understanding and enabling its use in applications like robotics, autonomous systems, and immersive virtual environments.

\section{Method}

In this work, we propose a novel approach to 3D scene understanding that facilitates object search in both image space and world space using open-vocabulary text queries. While previous methods\cite{langsplat,LEGaussians,feature3dgs} have achieved notable success in embedding language features into Gaussian Splatting via differentiable rendering for open-vocabulary segmentation in image space, they face significant challenges in 3D. Specifically, the reliance on alpha blending to learn semantics limits their ability to accurately capture and represent 3D semantics, hindering tasks such as object selection or extraction based on learned 3D features.

To address these limitations, our approach moves away from training individual Gaussians or pixels for language feature representation. Instead, we focus on grouping 3D Gaussians into distinct objects and associating language features with these abstract object groups rather than individual Gaussians. This shift enables open-vocabulary searches to operate at the object ID level, significantly simplifying tasks like 3D object extraction and 2D segmentation, as they reduce to filtering objects based on their assigned IDs.

The rest of this section is structured as follows: we first introduce the foundational concepts of the Gaussian Grouping\cite{gaussian_grouping} method, which forms the basis of our approach. Next, we discuss our Bag of Features Embedding strategy for representing grouped objects. Finally, we detail how we implement open-vocabulary text queries to support object search in both 2D and 3D spaces.

\subsection{Preliminaries: Gaussian Grouping}

Gaussian Grouping\cite{gaussian_grouping} introduces a method to enable fine-grained 3D scene understanding by assigning instance or stuff IDs to 3D Gaussians. This is achieved by augmenting Gaussians with a learnable Identity Encoding parameter, a compact vector of length 16, which represents unique object identities across the scene.

\subsubsection{Input Preparation}

We first generate 2D masks from multi-view images using the SAM model\cite{sam}. Each mask is assigned a unique ID across views by treating the multi-view setup as a video sequence and employing a zero-shot tracker\cite{deva} for mask consistency. The output from this stage is a consistent pixel-aligned object ID for each object in the image.

\subsubsection{3D Gaussian Rendering and Identity Encoding}

The Identity Encoding is optimized using a differentiable 3D Gaussian renderer. This renderer projects 3D Gaussians into 2D mask identity features by blending Identity Encoding vectors with Gaussian alpha contributions resulting in an output $I \in R^{16\times H\times W}$. Finally, a linear layer is applied followed by a softmax activation function to the rendered Identity Encoding of each pixel in the 2D image. This transforms the encoded features into a probability distribution, indicating the likelihood of each pixel belonging to a specific object ID in the scene.

\subsubsection{Grouping Loss}

To supervise identity learning, Gaussian Grouping combines two losses:

\begin{itemize}
    \item \textbf{2D Identity Loss}: Uses cross-entropy classification over the rendered 2D mask identities, ensuring alignment with the 2D input labels.
    \item \textbf{3D Regularization Loss}: Encourages spatial consistency by minimizing feature distances between neighboring Gaussians in 3D space. This ensures robust identity learning even for occluded or sparsely visible objects.
\end{itemize}

\begin{figure}[htp!]
    \centering
    \includegraphics[width=0.4\textwidth]{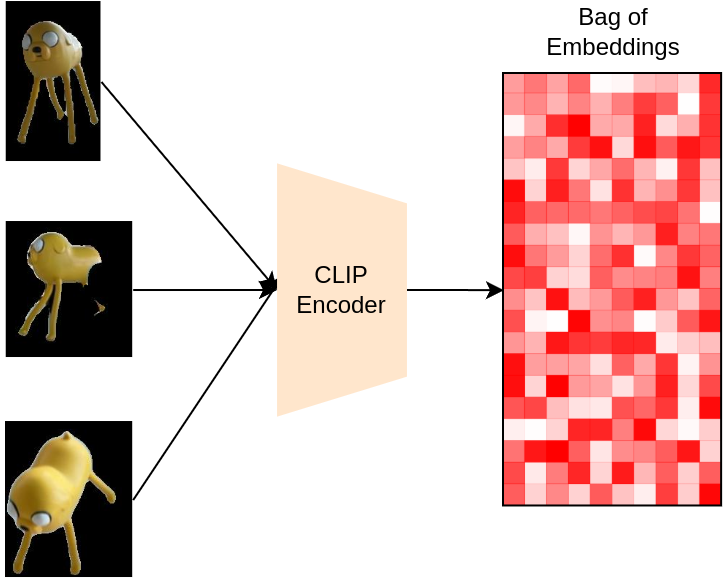}
    \caption{\textit{Multiview Bag of Features Embedding Extraction.} Given multiple views pf a masked 3D object, we extract the visual embeddings of each view using CLIP image encoder. These multiview embeddings are aggregated into a bag-of-embeddings representation, which captures the semantic content of the object from different perspectives. This bag-of-embeddings representation enables open-vocabulary 2D segmentation and 3D object selection using text queries.}
    \label{fig:bag_of_embeddings}
\end{figure}

\subsection{Multiview Bag of Features Embedding}

To enable open-vocabulary queries about fine-grained object attributes, we augment each 3D object with a bag of CLIP embeddings aggregated across all its 2D appearances. While a straightforward approach would involve averaging embeddings across views to create a unified representation, this method risks discarding semantic details visible only in specific perspectives, such as a book title visible only from a close-up view or a logo visible exclusively in frontal views. By preserving the full set of embeddings, our method retains these rare but distinctive characteristics, allowing language queries to target both prominent and subtle attributes.

The extraction of language embeddings operates independently from the training of Gaussian splatting with group IDs, which allows performing these two steps in parallel. The process begins with the input data prepared for Gaussian grouping \cite{gaussian_grouping}. For each object ID, the corresponding object is masked across all training views to isolate it from the background and other objects. As illustrated in Figure \ref{fig:bag_of_embeddings}, the masked views are passed through a CLIP encoder to extract semantic features for each view. These features are then aggregated into a bag of embeddings, enabling open-vocabulary searches for relevant object IDs based on text queries.

This strategy takes advantage of CLIP’s ability to retrieve fine-grained details. Since each embedding represents a specific view of the object, the collection of embeddings captures the full diversity of the object. During a query, this approach can match not only the common features of the object but also the unique details visible from specific angles. This ensures that the representation of an object is not limited to its "average" appearance but reflects all its visual traits, even those that appear infrequently across views.

\begin{figure}[htp!]
    \centering
    \includegraphics[width=0.8\textwidth]{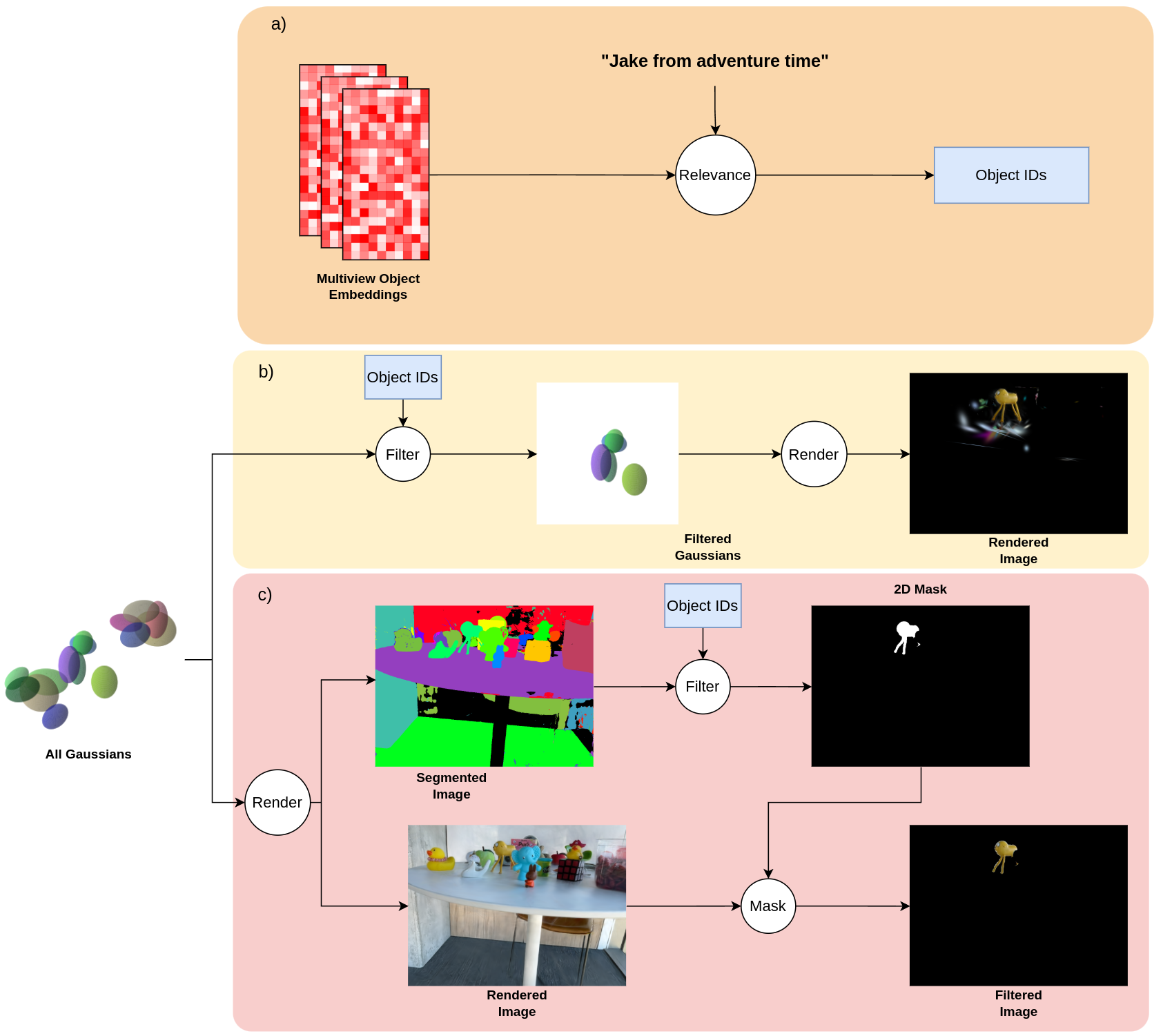}
    \caption{\textit{Overview of our open-vocabulary 3D object selection and 2D segmentation pipeline.}
    (a) Multiview image embeddings of 3D objects are matched with a text query (e.g., "\textit{Jake from adventure time}") to extract relevant object IDs.
(b) Using the retrieved object IDs, we filter the 3D Gaussians to render only the relevant object, enabling 3D object selection.
(c) To perform open-vocabulary 2D segmentation, the full set of Gaussians is rendered from a specific camera view to produce a photorealistic image and an accompanying segmentation map. The object IDs obtained in (a) are then used to generate a binary mask that localizes the queried object in the image, enabling text-driven segmentation directly in rendered views.}

    \label{fig:object_search}
\end{figure}

\subsection{Open-vocabulary Search}

As shown in Fig. \ref{fig:object_search}(a), we perform a text query search in the object embedding space to identify the object IDs most relevant to an open-vocabulary input text query. The task is formulated as follows: Given an object $i$ represented by a bag of embeddings $\phi_{object}^i$, which are extracted following the process illustrated in Fig. \ref{fig:bag_of_embeddings}, we compute the relevancy $\rho^{i,j}$ between each embedding $\phi_{object}^{i,j}$ and a text query $\phi_{query}$ . The relevancy score is calculated using the approach described in LERF \cite{lerf} and is defined as

\[\rho^{i,j} = \min_k \frac{\exp(\phi_{object}^{i,j}.\phi_{query})}{\exp(\phi_{object}^{i,j}.\phi_{query})+\exp(\phi_{object}^{i,j}.\phi_{canon}^k)}\]

Here, $\phi_{canon}^k$ represents the CLIP embeddings of predefined general canonical queries such as "object", "stuff", and "texture". This formulation computes the relevancy score between a single embedding of an object and the text query, effectively measuring how well the embedding aligns with the given query in the presence of canonical concepts.

To compute the overall relevancy score for an object, we get the relevancy scores of all its embeddings with the input query. The final score is derived by averaging the top $k$ relevancy scores, where $k$ is a hyperparameter that balances capturing fine-grained semantic details and ensuring accurate object identification. This process is mathematically expressed as:
\[\rho^i = \frac{1}{k} \sum_{j=1}^k \rho^{i,[j]}\]

where $\rho^{i,[j]}$ denotes the $j$-th highest relevancy score among all embeddings of object $i$.

With the object relevancy scores computed, searching in either image space or world space can be performed in a systematic manner. The object IDs with the highest relevancy scores are identified and their associated representations are extracted. For world-space object extraction, Gaussians corresponding to the selected object IDs are retrieved. For open-vocabulary segmentation in the image plane, pixels with matching IDs are masked to segment the relevant objects.

After extracting object IDs relevant to a text query, performing downstream tasks such as image segmentation or 3D object selection can be efficiently performed. As illustrated in Fig.\ref{fig:object_search}, filtering by object IDs is integrated at different stages of the rasterization process to accomplish the desired task. In panel b), the Gaussians are filtered based on their object IDs before rasterization, enabling 3D object selection. In panel c), the Gaussians are first rendered from the desired viewpoint, producing an RGB image along with a segmented image containing pixel-aligned object IDs. A mask is then generated from the segmented image using the extracted object IDs from the text query. Finally, this mask is applied to the rendered image to isolate and segment the desired object.

\section{Experiments}

\subsection{Setup}

\subsubsection{Datasets}
We evaluate our approach using two distinct datasets. The first dataset, LERF\cite{lerf}, contains various real-world scenes captured with a standard iPhone camera, providing a rich set of in-the-wild scenarios. The second dataset, 3D-OVS\cite{3dovs}, features a wide range of object categories, including those with long-tail distributions. To evaluate 2D mask segmentation on the 3D-OVS dataset, we follow the evaluation protocol specified in \cite{3dovs}. While for LERF dataset, we follow the evaluation protocol outlined in \cite{langsplat}.

\subsubsection{Metrics}
We evaluate the performance of our method on both
2D and 3D semantic segmentation tasks. For novel view
synthesis of 2D masks, we use the mean Intersection over Union (mIoU) and localization accuracy. For 3D object extraction, we present a qualitative and visual evaluation of our method compared to the state-of-the-art.

\subsubsection{Implementation Details}
We implement our method based on Gaussian Grouping\cite{gaussian_grouping} with the same hyperparameters provided in the paper with an identity encoding of 16 dimensions for each Gaussian. Moreover, during open-vocabulary search queries, we use $k=5$ for averaging the top relevancy score and we employ a postprocessing step by ignoring objects that were not visible in at least 20\% of the images in the dataset to remove the noisy output coming from DEVA\cite{deva}.

\begin{figure}[htp!]
    \centering
    \includegraphics[width=0.5\textwidth]{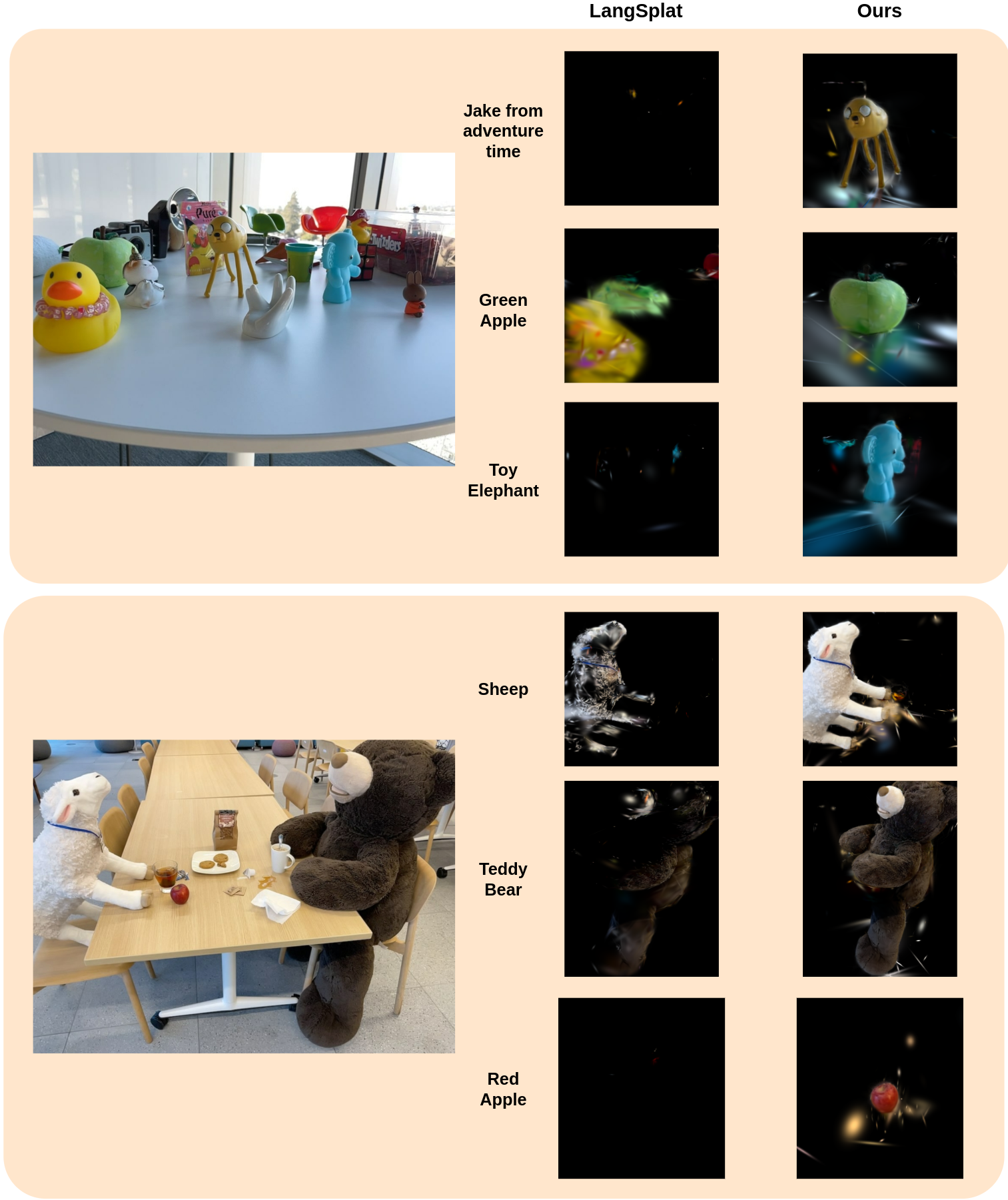}
    \caption{A qualitative comparison between our method and Langsplat\cite{langsplat} on open-vocabulary 3D object extraction on the LERF dataset\cite{lerf}. Given the same text queries, our method produces significantly more accurate object reconstructions, with the extracted Gaussians preserving both the geometry and appearance of the target object. In contrast, LangSplat often fails to recover complete objects—frequently producing fragmented or partial results, with large portions of the target object missing. This highlights the effectiveness of our approach in achieving precise and complete 3D object selection from text.}
    \label{fig:results}
\end{figure}

\subsection{Open Vocabulary 3D Object Selection}

Our primary focus in this work is on advancing 3D object extraction using open-vocabulary text queries. Traditional methods for embedding semantics into Gaussians often depend on explicit language training, which can be prone to noise, especially when using differentiable rendering. In contrast, our approach addresses this limitation by leveraging the strengths of Gaussian Grouping \cite{gaussian_grouping} and multiview semantic embeddings on the object level, enabling robust and flexible open-vocabulary 3D object extraction.

The process illustrated in Fig.\ref{fig:object_search}b begins with a text query, which retrieves the most relevant object IDs based on semantic similarity. Once these IDs are identified, the Gaussians associated with them are extracted, facilitating a structured and meaningful reconstruction of the objects in 3D space. By operating directly on semantic representations, our method bypasses the challenges of explicit language training, ensuring greater accuracy and resilience to noise.

To evaluate the effectiveness of our approach, we performed a qualitative comparison against state-of-the-art methods, as shown in Fig.\ref{fig:results}. Our method demonstrates the ability to extract Gaussians representing objects solely based on input text queries, while preserving both the geometry and appearance of the objects. For instance, in the "green apple" text query, our approach reconstructs the complete geometry of the object from the desired view, even when it is largely occluded in the original view. This highlights the advantages of 3D object extraction in uncovering occluded structures and ensuring a comprehensive representation. In contrast, LangSplat \cite{langsplat}, while effective for 2D image segmentation, struggles with 3D text queries, often producing noisy or incomplete outputs for the desired object. These comparisons underscore the robustness and versatility of our method in open-vocabulary 3D object extraction.

\subsection{Open Vocabulary Querying in 2D}

Although the primary objective of this work is 3D object extraction, we demonstrate that our approach also delivers competitive performance in open-vocabulary 2D segmentation, aligning with state-of-the-art methods and even surpassing them in certain cases. This underscores the versatility and robustness of our method in addressing both 3D and 2D tasks.


\begin{table}[h!]
\setlength{\tabcolsep}{6pt} 
\centering
\begin{tabular}{lcccc}
\toprule
\textbf{Methods} & \textbf{figurines} & \textbf{ramen} & \textbf{teatime} & \textbf{overall}\\
\midrule
LERF\cite{lerf}          & 28.2 & 37.9 & 38.6 & 37.3 \\
Langsplat\cite{langsplat}      & 44.7 & 51.2 & 65.1 & 53.7\\
Gaussian Grouping\cite{gaussian_grouping}        & \textbf{69.7} & 45.8 & 61.9 & 59.1\\
\midrule
Ours                & 63.3 & \textbf{53.7} & \textbf{79.3} & \textbf{65.4}\\
\bottomrule
\end{tabular}
\caption{IOU performance on LERF dataset\cite{lerf}. The best performance per column across all methods is indicated in \textbf{bold}.}
\label{tab:lerf_iou}
\end{table}


\begin{table}[h!]
\setlength{\tabcolsep}{6pt} 
\centering
\begin{tabular}{lcccc}
\toprule
\textbf{Methods} & \textbf{figurines} & \textbf{ramen} & \textbf{teatime} & \textbf{overall}\\
\midrule
LERF\cite{lerf}          & 75.0 & 62.0 & 84.8 & 58.1 \\
Langsplat\cite{langsplat}      & 80.4 & 73.2 & 88.1 & 80.6\\

\midrule
Ours                & \textbf{87.5} & \textbf{74.7} & \textbf{100} & \textbf{87.4}\\
\bottomrule
\end{tabular}
\caption{Localization Accuracy performance on LERF dataset\cite{lerf}. The best performance per column across all methods is indicated in \textbf{bold}.}
\label{tab:lerf_accuracy}
\end{table}


\begin{table}[h!]
\setlength{\tabcolsep}{6pt} 
\centering
\begin{tabular}{lcccccc}
\toprule
\textbf{Methods} & \textbf{bed} & \textbf{bench} & \textbf{room} & \textbf{lawn} & \textbf{sofa} & \textbf{overall}\\
\midrule
LERF\cite{lerf}          & 88.9 & 71.4 & 66.1 & 81.2 & 77.5 & 77 \\
3D-OVS\cite{3dovs}          & 89.5 & 89.3 & \cellcolor{orange!20}92.8 & 88.2 & 74 & 86.8 \\
LEGaussians\cite{LEGaussians}          & 45.7 & 47.4 & 44.7 & 49.7 & 48.2 & 47.1 \\
Langsplat\cite{langsplat}      & \cellcolor{orange!20}92.5 & \cellcolor{orange!20}94.2 & \cellcolor{green!25}94.1 & \cellcolor{green!25}96.1 & \cellcolor{green!25}90 & \cellcolor{green!25}93.4\\

\midrule
Ours                & \cellcolor{green!25}98 & \cellcolor{green!25}96.1 & 89.7 & \cellcolor{orange!20}95 & \cellcolor{orange!20}78.2 & \cellcolor{orange!20}91.4\\
\bottomrule
\end{tabular}
\caption{IOU performance on 3D-OVS dataset\cite{3dovs}. Best values are highlighted in \fcolorbox{white}{green!25}{green}, while second best values are highlighted in \fcolorbox{white}{orange!20}{orange}.}
\label{tab:3dovs_iou}
\end{table}

The experimental setup for open-vocabulary 2D segmentation, illustrated in Fig.\ref{fig:object_search}c), involves rendering 3D Gaussians into a specific 2D view. Each Gaussian is rendered along with its identity encoding, producing a pixel-wise categorization of object IDs. Using an input text query, we identify the objects most semantically relevant to the query. These relevant object IDs are then used to segment the 2D view by isolating the pixels corresponding to these IDs.

As shown in Tables \ref{tab:lerf_iou} and \ref{tab:lerf_accuracy}, our model, evaluated on the LERF dataset \cite{lerf}, surpasses the state-of-the-art methods, including LangSplat \cite{langsplat}, across both metrics, demonstrating the effectiveness of our multiview bag of embeddings representation. Additionally, Table \ref{tab:3dovs_iou} presents our performance on the 3D-OVS dataset \cite{3dovs}, where our method outperforms the state-of-the-art on two individual datasets and ranks second overall. The slight drop in overall performance is attributed to certain queries in this dataset, which involve segmenting background regions. These scenarios pose challenges for the DEVA model \cite{deva}, which sometimes struggles to fully capture background semantics, making it more difficult for the Gaussians to accurately learn the background identity.


\section{Conclusion}

In this work, we introduced a novel approach to 3D scene understanding that addresses the limitations of existing methods in open-vocabulary object retrieval and segmentation. By utilizing predecomposed object-level Gaussians from Gaussian Grouping and representing each object with a robust bag of multiview CLIP embeddings, our method achieves accurate and reliable open-vocabulary search in both 2D and 3D spaces. Unlike prior approaches that depend on differentiable rendering for semantic learning—often resulting in noisy and inconsistent per-Gaussian features—our method circumvents these challenges by directly operating on object-level semantics, ensuring a cleaner and more coherent representation.

Our experimental results demonstrate that the proposed method excels in 3D object extraction while achieving competitive or superior performance in open-vocabulary 2D segmentation. The ability to seamlessly propagate object-level semantic queries to both pixel-level segmentation and 3D Gaussian selection underscores the flexibility and adaptability of our framework. Notably, the method effectively reconstructs occluded structures and supports fine-grained 3D object selection, further solidifying its utility across diverse applications.

Future directions for this research include integrating advanced vision-language models to enhance semantic understanding, improving the robustness of object grouping in complex or cluttered scenes, and extending the framework to dynamic scenes with temporal consistency.

Ultimately, our method represents a significant step forward in bridging the gap between high-fidelity rendering and structured 3D scene understanding. By enabling robust open-vocabulary search and segmentation in 2D and 3D, it opens new avenues for applications in AR/VR, robotics, and interactive 3D editing, paving the way for more intuitive and scalable scene comprehension.

%

%
%
\bibliographystyle{plain}
\bibliography{bibexample}
\end{document}